\pdfoutput=1

\documentclass[11pt]{article}

\usepackage{acl2023} 

\usepackage{times}
\usepackage{latexsym}
\usepackage{graphicx}
\newcommand{\name}{Go-tuning }

\usepackage[T1]{fontenc}

\usepackage[utf8]{inputenc}

\usepackage{microtype}
\usepackage{algorithm}
\usepackage{amsmath}
\usepackage{times}
\usepackage{latexsym}
\usepackage{amsmath,amsfonts,amssymb,bbm,epsfig,bm}
\usepackage{graphicx}
\usepackage{booktabs}
\usepackage{multirow}
\usepackage{makecell}
\usepackage{subfigure}
\usepackage{framed}
\usepackage{pstricks}
\usepackage{xcolor}
\usepackage{color}
\usepackage{enumerate}
\usepackage{arydshln}
\usepackage{algorithm}
\usepackage{longtable}
\usepackage{ltablex}
\usepackage{array}
\usepackage{makecell}
\title{Go-tuning: \\ Improving Zero-shot Learning Abilities of Smaller Language Models}

\author{
Jingjing Xu\textsuperscript{\rm1},
Qingxiu Dong\textsuperscript{\rm2},
Hongyi Liu\textsuperscript{\rm3}, and Lei Li\textsuperscript{\rm4} \\
\textsuperscript{\rm 1}Shanghai AI Lab \\
\textsuperscript{\rm 2} MOE Key Lab of Computational Linguistics, School of Computer Science, Peking University \\
  \textsuperscript{\rm 3} Shanghai Jiao Tong University,
  \textsuperscript{\rm 4} University of California, Santa Barbara \\
  \texttt{jingjingxupku.02@gmail.com},\texttt{dqx@stu.pku.edu.cn} \\
  \texttt{liu.hong.yi@sjtu.edu.cn},\texttt{
  lilei@cs.ucsb.edu }
}

\begin{document}

\maketitle

\begin{abstract}
With increasing scale, large language models demonstrate both quantitative improvement and new qualitative capabilities, especially as zero-shot learners, like GPT-3. However, these results rely heavily on delicate prompt design and large computation.
In this work, we explore whether the strong zero-shot ability could be achieved at a smaller model scale without any external supervised data. 
To achieve this goal, we revisit masked language modeling and present a geometry-guided self-supervised learning method (\name for short) by taking a small number of task-aware self-supervised data to update language models further. Experiments show that \name can enable T5-small (80M) competitive zero-shot results compared with large language models, such as T5-XL (3B). We also apply \name on multi-task settings and develop a multi-task model, mgo-T5 (250M). It can reach the average performance of OPT (175B) on 9 datasets. 
\end{abstract}

\section{Introduction}


It has been well observed that large language models (LLMs) manifest extraordinary zero-shot and few-shot performance~\citep{DBLP:journals/corr/abs-2005-14165,DBLP:journals/corr/abs-2206-07682,DBLP:conf/iclr/WeiBZGYLDDL22}. 
With the increasing scales, large models make it possible to handle various tasks by simply conditioning the models on instructions describing the task~\citep{DBLP:conf/emnlp/PetroniRRLBWM19,DBLP:journals/corr/abs-2005-14165}, prompt for short, without training on supervised data. 
And such surprising zero-shot performance has been the key milestone of recent language model revolutions.  



\begin{figure}[t]
    \centering
    \includegraphics[width=\linewidth]{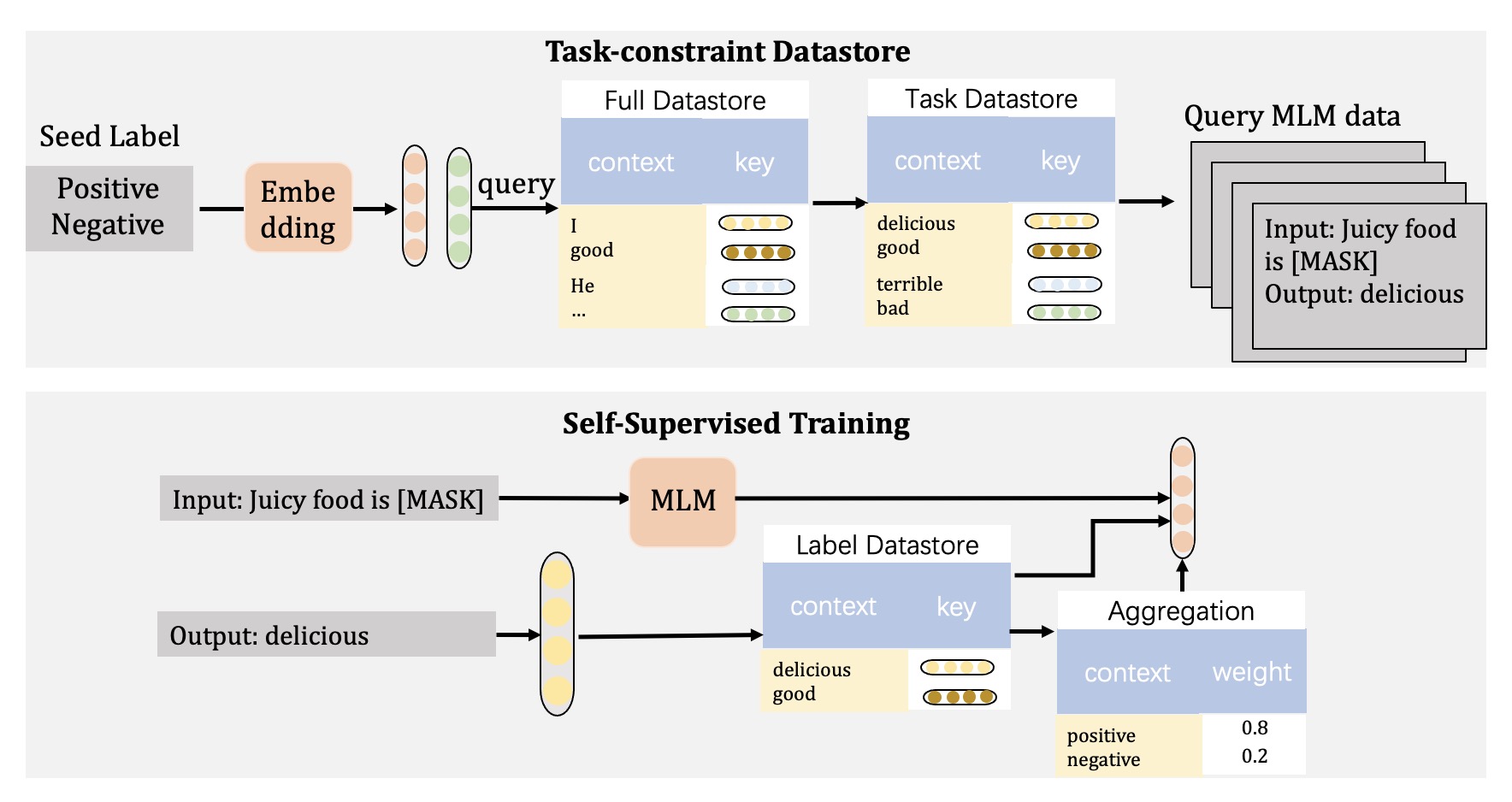}
    \caption{Illustration of Go-tuning. \textbf{The upper part} shows the process of extracting self-supervised data. we first build a symbolic datastore storing all tokens associated with representations (full datastore). For an understanding task, we suppose the seed label is given. We use the seed labels to extract task-aware  sub-datastore (task datastore). The task datastore is used to extract MLM training data. \textbf{The lower part} shows self-supervised training. We apply kernel density estimation to aggregate the probability of related tokens together.   }
    \label{fig:go}
\end{figure}

Despite excellent results, the strong zero-shot abilities are highly dependent on LLMs~\citep{DBLP:journals/corr/abs-2206-07682} and are sensitive to the prompt surface~\citep{DBLP:conf/emnlp/PetroniRRLBWM19,calibrate}. 
The major challenge for zero-shot inference lies in the gap between the pre-trained landscape over millions of tokens and the task-constraint landscape over label tokens. To bridge the gap, prompt-tuning has been a common solution~\cite{DBLP:journals/corr/abs-2109-03564,DBLP:journals/corr/abs-2206-07682}. The prompt-tuning method induces LLMs to generate a prediction distribution similar to task-aware prediction distribution, where the right label tokens have higher probabilities, such as \textsl{``Juicy beef is a [positive/negative] review''}. However, such estimation is a ``cherry-pick'' game and largely depends on the ability of LLMs. 


In this work, we aim to extrapolate the zero-shot ability of language models, especially for small-size and normal-size language models.  There is a multi-task research direction borrowing supervised data from other tasks to boost the zero-shot performance of smaller language models, like T0~\citep{DBLP:journals/jmlr/RaffelSRLNMZLL20} and FLAN-T5~\citep{DBLP:journals/corr/abs-2210-11416}. In this work, we do not suppose large-scale supervised data and task similarity; we explore a more challenging but task-agnostic setting: can smaller language models learn strong zero-shot abilities just from self-supervised data? 



To this end, we revisit masked language modeling (MLM) and propose a geometry-guided self-tuning method, Go-tuning, to smoothly induce smaller models to transfer from a pre-trained prediction landscape to a task-constrained prediction landscape. 
The key idea of Go-tuning is quite simple: selecting task-aware MLM training data to 
train the models to learn task-constraint token predictions. The whole landscape transfer process only relies on a small number of self-supervised data and can easily extend to various tasks.  

Given a pre-trained language model, we first build a symbolic datastore storing all tokens associated with representations (embeddings). We suppose the seed label and a task prompt are given for an understanding task. 
Starting from seed labels, we query the datastore to extract the nearest label tokens according to the symbolic datastore. We then train the small models on task-aware MLM training data with the extracted tokens as predictions.
During training, we consider the distance between the extracted labels and seed labels and adopt kernel density estimation to complete a unified downstream task
 


Despite the simplicity, \name successfully
improves zero-shot performance of small-size and normal-size models. Importantly, \name is versatile and task-agnostic and can facilitate understanding across various tasks even cross-task learning. 
In addition, \name also shows a strong ability to extrapolate multi-task zero-shot learning abilities. We take T5 as the backbone and develop multi-task version T5, mgo-T5, with 250M parameters. Experiments show that mgo-T5 outperforms larger language models, such as T5-3B and even reaches the average performance of OPT-175B. 
While the vibrant field of LLMs started out from the premise of excellent few-shot learners, we hope our work may encourage more research into zero-shot capabilities hidden inside smaller models.

\section{Background}
We briefly review the related studies: large language models and zero-shot prompting, and improving zero-shot abilities of language models. 

\paragraph{Large language models and zero-shot prompting}
Basically, current LLMs usually adopt prompting to induce models to extrapolate the zero-shot abilities. The key idea is to imitate the pre-training data distribution and create a masked natural language text, like \textsl{"Juicy beef is a [MASK] review."}. By filling out candidate answers and the final answer is selected with a lower perplexity score. Since such zero-shot learning abilities are positively related to model size (associated with data size), these abilities are also regarded as the emergent abilities of large language models. In addition to template-based prompts, there are also instruction-based prompts. ~\citep{DBLP:journals/corr/abs-2205-11916} showed that LLMs are decent zero-shot reasoners by simply adding ``Let's think step by ste'' before each answer. ~\citet{DBLP:journals/corr/abs-2203-02155} built a more real-world way to teach models to complete a specific task. like ``Write a short story where a bear goes to the beach, makes friends with a seal, and then returns home‘’.



\paragraph{Improving zero-shot learning abilities} 

Instruction fine-tuning is a hot research direction to improve the zero-shot abilities of LLMs by training models on a collection of tasks phrased as instructions~\citep{DBLP:journals/corr/abs-2203-02155}. ~\citet{DBLP:journals/corr/abs-2203-02155} aligned GPT with user intent on a wide range of tasks by fine-tuning with human feedback.  ~\citet{DBLP:conf/iclr/WeiBZGYLDDL22} proposed a Flan-T5 model. They find that instruction fine-tuning does scale well with the number of tasks and the size of the model. 
Following the chain-of-though research direction, ~\citet{DBLP:journals/corr/abs-2210-11610} explored zero-shot question-answering tasks. They used the answers generated by a pre-trained LLM based on chain-of-thought prompting as pseudo labels and augmented these data to fine-tune LLMs.     Unlike these studies, we do not suppose large-scale supervised data or large-scale human-written instructions, we explore a more challenging question: Whether smaller models can learn zero-shot abilities just from self-supervised data.

\begin{table*}[t]
\footnotesize
\setlength{\tabcolsep}{4pt}
\centering
\resizebox{0.99\textwidth}{!}{
    \begin{tabular}{@{}lm{2cm}<{\centering}m{12cm}<{}}
    \toprule
    \textbf{Task} & \textbf{Seed Label} &\textbf{Neighbor Label Sets}  \\
    \midrule
        Sentiment Classification& positive, negative & good, bad, positive, negative, negatives, positives, Gooding, positively, negatively, nice, terrible, worst, excellent, decent, horrible, favorable, goodie, excellently, worse, badly, awful, satisfactory, evil, decently, goodness, poorly, GREAT, goodies, pleasant, Terrific, worsen, favourable, rotten, glad, Nicely, nicest, splendid, enjoyable, disastrous, best, pleased, fine, smoothly, worsening, WELL, nasty, greatness, detrimental, favorably, satisfying, affirmative, worsened, upbeat, badass, dreadful, shame, praise, praised, pleasing, sour, evils, beneficial, relieved, minus, horribly, crappy, unpleasant, thankful, praising, cheerful, deteriorating, optimistic, miserable, satisfaction, downfall, praises, happiness, Blessings, negativity, ruin, pleasantly, reassuring, misery, shameful, Poor, benefited, weakness, plagued, optimism, filthy, harm, Blessing, satisfied, adversely, grateful, decreases, HAPPY, decrease, loss, disapproval \\

        \midrule
        Natural Language Inference & entailment, neutral, contradiction &  entail, entailed, entail, implies, imply, implication, entails, inferences, entailed, predicate, inference, assertion, injunctive, assertions, consequent, conjunction, excludes, correctness, omission, paraphrase, asserts, implication, clarifying, contradictions, logically, omissions, justifying, statements, Statements, clauses, comprehension, exclusion, implications, Implications, semantic, Semantic, theorem, complements, implies, conclusion, Conclusion, injunctive, accusation, hypotheses, relevancy, semantics, quantification, predicate, conclusions, Conclusions, logic, Logic, exclude, Clause, clause, contradict, juxtaposition, inferences, contradicts, inference, declarations, associating, conditional, entails, causation, imply, exclusions, refute, indicates, interpreting, disqualify, avoids, avoiding, Avoiding, negate, logically, denies, assertions, contradiction, denying, misleading, refusal, deny, paradox, Paradox, preclude, avoidance, paradoxical, contradictory, clause, Clause, dichotomy, hypotheses, counterpoint, relevancy, comprehension, conditional, unambiguous, semantic \\
  \bottomrule
    \end{tabular}
    }
    \caption{The extracted neighbor label sets. We use the seed labels to query the label datastore to build a task-aware label datastore. Here is some examples constructed for sentiment classification and natural language inference.    }
    \label{tab:my_label}
\end{table*}

\section{Approach}

We propose \name a self-supervised method to extrapolate zero-shot abilities of language models. 
It differs from current prompt-based zero-shot learners as it does not require delicate human-designed prompts, and it differs from instruction-finetuning-based zero-shot learners as it does not rely on supervised training over multiple NLP tasks and human-written prompts. The core idea of our method is simple: adding a task-aware self-supervised training stage before inference. For simplification, we take T5 as the backbone and tune it to our methods.   



The whole approach contains three steps: 1) build a symbolic label-space datastore recording all prediction tokens associated with representations; 2) given a specific task with seed labels, query datastore to extract neighbor labels, which are then used to extract task-aware MLM training data; 3) constrained training to learn task-aware prediction landscape. 

\subsection{Building Label-space Datastore}
Since downstream tasks only contain limited labels and the first question is to transfer a pre-trained model to task-aware models. We first implement a language model to build a symbolic datastore recording all label tokens and their embeddings. For a token in the dictionary, we adopt its embedding representation used for the last decoder layer for prediction as keys and its symbolic surface as value. The datastore $D$ consists of key-value pairs.

\subsection{Extracting Self-supervised Data}


For a task, we suppose that the seed label is given. \name then extracts the nearest neighbor labels to augment self-supervised training data. 

\paragraph{Neighbor Sets.} Given a specific task, we use label embedding to identify neighbor tokens based on datastore $D$. The neighbor tokens are selected via label similarity.   For any training data: Given y, this step can be formulated as the problem that finds the nearest label $z_1, \cdots, z_n$. We adopt the dot-product distance following pre-trained targets. 
 \begin{equation}
     d(y,z_i) = y \cdot  z_i
 \end{equation}
 where $y$ is one label of specific tasks. 
 Table~\ref{tab:my_label} shows some cases about the extracted neighbor labels according to task-specific labels. 
 Given neighbor labels, we extract MLM training data with extracted label tokens as predictions for further fine-tuning. 
 

\subsection{Constrained Training}

Given a set of auxiliary neighbor labels, this part shows how to use the auxiliary data to help the training of target datasets.

 \paragraph{Neighbor Density Estimation} 
 The nearest labels can avoid the effects of the label surface on the model performance and thus reduce label engineering on designing specific labels. However, the introduced label tokens are not always have the exact same meanings, like \textsl{``negative''} and \textsl{``ruin''}.  To encourage models to learn a unified task, we consider the distance between these label tokens by using density estimation. In this way, the probability of $p(y|x)$ can be represented as: 
\begin{align}
\centering
    p(y|x) &= \sum_{z} p(y, z|x) \\
  &\approx  \sum_{z_i}\frac{exp(y,z_i)}{\sum_{k}exp(z_k,y)}p(z_i|x) 
  \label{eq:obj}
\end{align}
where $z={z_1, \cdots, z_n}$ represents the neighbor labels. 
Due to the lack of all labels for a single input, we can not directly optimize Eq.~\ref{eq:obj} via sample-based methods. Therefore, we apply the following bi-level optimization algorithm:
First, we propose to optimize the separated distribution: 
$$p(y|x), w_1p(z_1|x),..., w_tp(z_t|x)$$ 

where $$w_t = w(y, z_t, \phi)$$ and $\phi$ are randomly initialized to learn the separated distribution. Then, we use $$|| p(y|x) - w_1p(z_1|x) - ... - w_tp(z_t|x)||$$ to learn $w(y, z_t, \phi)$ and constraint the mass of $p(y|x)$ equal to the estimated mass. 

The constrained training only requires label attributes and does not require any annotated labels. The training data come from self-supervised data with $y$, $z_1,\cdots,z_t$ as labels.



\begin{table*}[t]
\footnotesize
\setlength{\tabcolsep}{2pt}
\centering
    \begin{tabular}{@{}llm{4cm}<{\centering}m{5cm}<{\centering}}
    \toprule
   \textbf{Dataset} & \textbf{Task}  & \textbf{Seed label} & \textbf{Template}  \\
    \midrule
        SST2 & Sentiment Classification & positive, negative & Sentiment of the review [input] is [label] \\
        \midrule
        QNLI, RTE, CB,WNLI & Natural Language Inference & entailment, neutral, contradiction & Given [input1], it is [label] that [input2] happens \\
          \midrule
        MRPC,PAWS & Paraphrasing & paraphrase, non-paraphrase & Sentence1 [entity1] and sentence2 [entity2] have the relation of [label] \\
          \midrule
        BOOLQ & Question Answering & true, false & Given [entity1], [entity2] is a [label] statement \\
          \midrule
        COLA & Grammar Checking & unacceptable, acceptable & Grammar of sentence [entity1] is [label] \\
        
  \bottomrule
    \end{tabular}
    \caption{The evaluation benchmark, covering 5 tasks and 9 datasets. The template and seed labels are pre-defined. All baselines share the same template and seed labels. }
    \label{tab:task}
\end{table*}

 \section{Self-supervised Multi-task Learner}
Since \name aims at modeling label relations, it is natural to apply it directly on multi-task settings as a self-supervised multi-task learner. Furthermore, the multi-task settings also can address each-task-each-model problems in the proposed approach. For geometric-guided learning, we build multi-task training data without any labeled data. To be specific, we collect 260 tasks associated with human-written prompts and seed labels.  We first merge seed labels across different tasks together and query the MLM raw text to extract multi-task self-supervised training data.  The self-supervised data contains self-supervised training pairs with task-specific label attributes as labels.

\section{Experiments}

\subsection{Tasks and Datasets}
We evaluate the proposed methods on five kinds of understanding tasks: sentiment classification, natural language inference, question answering, grammar checking, and paraphrasing. as shown in Table~\ref{tab:task}. Following ~\citet{DBLP:conf/iclr/WeiBZGYLDDL22}, we report the test accuracy if a test set is provided by Huggingface datasets\footnote{https://huggingface.co/datasets}; otherwise, we use the validation set as our test set.

\name requires self-supervised data for fine-tuning backbone models. We use the extracted label tokens to query MLM training text and extract MLM data with the extracted label tokens as predictions. Specifically, we adopt a subset of C4\footnote{https://huggingface.co/datasets/c4} as a raw text source. For simplification, we set the maximum  number of the extracted sentences to be 100,000 for single-task fine-tuning, to be 200,00 for multi-task fine-tuning. We will release all our data, including the extracted and raw texts. 
\begin{itemize}
 \item SST2. SST2 is a sentiment classification dataset. It contains reviews and human-written sentiment labels. It uses a two-way (positive/negative) class split, with only sentence-level labels.
 \item COLA. COLA is a dataset for grammar check. Each example is a sentence with human-written labels deciding whether it is a grammatical sentence.
 \item QNLI. QNLI is a question-paragraph task. The task is to determine whether the context sentence contains the answer to the question.  
 \item RTE. RTE is a natural language inference dataset with two-way labels, including entailment, and not-entailment.
 \item CB.  CB is also a natural language inference dataset. 
 \item BOOLQ. BOOLQ is a question-answering dataset for yes/no questions containing 15,942 examples. These questions are usually naturally occurring.
\item WNLI. WNLI is a natural language inference dataset. The model needs to predict whether the original sentence entails the sentence with the pronoun substituted. 
\item PAWS. PAWS is a paraphrasing dataset. It follows two-way labels. There are two possible labels: $0$ indicates that the pair has different meanings, while $1$ indicates that the pair is a paraphrase.

\end{itemize}

\begin{table*}[t]
\footnotesize
\setlength{\tabcolsep}{5pt}
\centering
\begin{tabular}{llllllllllll}
\toprule
\textbf{Model} & \textbf{Param.} & \textbf{SST2} & \textbf{COLA} & \textbf{QNLI} & \textbf{RTE} &  \textbf{CB} & \textbf{BOOLQ} & \textbf{WNLI}   & \textbf{MRPC}  & \textbf{PAWS} & \textbf{AVG.} \\
\midrule
T5-base  & 250M  & 64.1 & 67.6 & 49.4 &52.7 & 41.0& 46.7 & 43.6 & 33.6   & 45.2 & 49.3 \\
 T5-3B & 3B & 66.2 &  51.6   & \textbf{51.4} & 51.2 & 28.5 & 53.4 &46.4 & 46.6  & 52.0  & 53.3\\
 \midrule
 T5-base-SDA & 250M & 58.2 & 62.5 & 49.0 & 49.8 & 46.4 & 61.3 & 42.2  & 65.2 & 54.5 & 52.6 \\
 
 \midrule
 go-T5-small & 80M  & 57.7 &  67.2  & 49.4 &  52.7 & 37.5 & 61.9 & \textbf{46.4} & 65.5 & \textbf{54.6} & 54.8  \\
 go-T5-base & 250M $\times$ 9 & 64.1  & 66.8  & 49.4 & 52.7 & 46.4 & 61.0 & \textbf{46.4} & \textbf{67.0}   & 54.4  & 56.5 \\
 go-T5-3B & 3B $\times$ 9 & 64.3  & \textbf{68.5}  & 50.1 & \textbf{53.0} & 32.1 & \textbf{62.0} & \textbf{46.4} &65.9  &  54.5 &  55.2 \\
  \midrule
mgo-T5-base & 250M & \textbf{75.4} & 68.4 & 49.5 & 52.3  & \textbf{57.1}  & 61.8 & \textbf{46.4} & 58.8   & 54.5 & \textbf{58.2} \\


\bottomrule
\end{tabular}
\caption{Zero-shot results of the proposed approach compared to T5 baselines. }
\label{tab:main}
\end{table*}



\begin{table*}[t]
\footnotesize
\centering
\setlength{\tabcolsep}{5pt}
\begin{tabular}{llllllllllll}
\toprule
\textbf{Model} & \textbf{Param.} & \textbf{COLA} & \textbf{MRPC} & \textbf{QNLI} & \textbf{RTE} & \textbf{SST2} & \textbf{CB} & \textbf{BOOLQ} & \textbf{WNLI}  & \textbf{PAWS} & \textbf{AVG.} \\
\midrule

 GPT-3 & 175B & - & - & - & 63.5 & 71.6 & 46.4 & 60.5  & - & - &  -  \\
 LaMDA-PT & 137B & - &- &  50.6 & \textbf{73.3} & 51.0 & 42.9 & \textbf{83.0} & \textbf{56.3}  & 45.5 &- \\
 GLaM & 64B & - & - & - & 68.8 & -&  33.9& \textbf{83.0} & - & -  \\
 \midrule
 OPT & 175B & \textbf{73.4} & \textbf{68.4}  & \textbf{51.2} & 48.3 & \textbf{91.9} & 28.5 & 43.0 &  47.8  & \textbf{54.5} & 56.3 \\
 \midrule
 go-T5 (base) & 250M $\times$ 9 & 66.8 & 67.0 & 49.4 & 52.7 & 64.1 & 46.4 & 61.0 & 46.4   & 54.4  & 56.5 \\
 mgo-T5 (base) & 250M & 68.4 & 58.8 & 49.5 & 52.3 & 75.4 & \textbf{57.1}  & 61.8 & 46.4 & \textbf{54.5} & \textbf{58.2} \\

\bottomrule
\end{tabular}
\caption{Zero-shot performance of \name compared to GPT-3, LaMDA-PT, GLaM, and OPT. ``Param.'' represents the sum of parameters during inference. The models are not shared in go-T5 models. $ \times 9$ means $9$ models in total. ``AVG.'' represents the average performance of 9 tasks.  The multi-task model mgo-T5  with only 0.14\% parameters achieves competitive results compared to OPT (175B). }
\label{tab:com_with_LLM}
\end{table*}

\subsection{Evaluation Details}

For each dataset, we pre-define related templates (prompts) and seed labels, as shown in Table~\ref{tab:task}. Different prompts may result in different zero-shot performances. To reduce the variance caused by prompt surface, we do not adopt prompt engineering, and all baselines share the same templates. 

During inference, we fill the template with seed labels and calculate its probability according to a backbone model. The seed label with a higher probability is the final answer. 

\subsection{Baselines}
The proposed methods do not suppose the backbone architecture. In this paper, we choose T5~\citep{DBLP:journals/jmlr/RaffelSRLNMZLL20} as the backbone for simplification.

To fairly show the effectiveness of the proposed approach, we implement the following methods. 

\begin{itemize}
     \item T5 variants~\citep{DBLP:journals/jmlr/RaffelSRLNMZLL20}. The proposed approach is built upon T5 variants, including T5-small, T5-base, and T5-3B. We report the zero-shot results of T5 variants with different sizes.
     \item go-T5. We apply \name on T5 variants to see how the proposed method performs on different models with different sizes. 
       \item T5-base-SDA. We also conduct an ablation study to see how geometric training contributes to the final performance. This baseline directly fine-tunes backbone models on the extracted data without geometric regularization.
   
    \item mgo-T5. We apply \name on multi-task settings and report the performance. All tasks share the same parameters.

\end{itemize}

\subsection{Results}

\begin{table*}[t]
\small
\centering
\setlength{\tabcolsep}{6pt}
\begin{tabular}{llllll}
\toprule
\textbf{Task} & \textbf{T5-3B} &  \textbf{mgo-T5} & \textbf{Task} & \textbf{T5-3B}  &  \textbf{mgo-T5} \\
& (3B) & (250M) & & (3B) & (250M) \\
\midrule
assin (full) ~\citep{fonseca2016assin}   & 22.4  & 65.5 & assin (ptbr)~\citep{fonseca2016assin}  & 19.6 & 67.6 \\ 
assin (ptpt)~\citep{fonseca2016assin} & 25.2 &  63.4 & assin2~\citep{real2020assin} & 57.6 & 49.6 \\
catalonia\_independence (catalan) & 35.8 & 40.2 & cdsc (cdsc-e) &92.6 &  92.4 \\
clue (afqmc)~\citep{xu-etal-2020-clue}  & 31.0 & 31.0  & clue (cnli)~\citep{xu-etal-2020-clue}  &33.6 & 32.1 \\
emotion~\citep{saravia-etal-2018-carer} & 14.4 & 39.2  & gnad10~\citep{Schabus2017}  & 0.59 & 29.3\\

\makecell[l]{guardian\_authorship (cross\_genre\_2) \\ \citep{article}} & 40.4 & 34.8 & \makecell[l]{guardian\_authorship (cross\_genre\_3) \\ \citep{article}} & 45.4 & 46.4 \\
\makecell[l]{guardian\_authorship (cross\_genre\_4) \\ \citep{article}} & 39.8 & 40.5 & \makecell[l]{guardian\_authorship (cross\_topic\_10) \\ \citep{article}} & 59.2 & 61.2 \\
\makecell[l]{guardian\_authorship (cross\_topic\_11) \\ \citep{article}} & 51.9 & 51.5 & \makecell[l]{guardian\_authorship (cross\_topic\_12) \\ \citep{article}} & 68.3 & 71.8 \\ 
hate\_speech\_filipino & 46.7 & 45.6  & hausa\_voa\_topics & 22.8 &  26.5 \\
 pragmeval (emergent) & 17.4 & 36.7  & pragmeval (emobank-arousal) & 50.6 & 52.1 \\
pragmeval (emobank-dominance) & 37.1 & 39.9 & pragmeval (emobank-valence) & 44.0 & 48.9 \\
pragmeval (persuasiveness-claimtype) &  68.4 & 31.5 &
pragmeval (persuasiveness-eloquence) & 24.4 & 42.2 \\
pragmeval (persuasiveness-relevance) & 67.8 & 48.8 &
ro\_sent~\citep{dumitrescu2020birth} & 45.6 & 43.9  \\
\makecell[l]{russian\_super\_glue (rcb) \\ \citep{shavrina2020russiansuperglue}} & 30.9 & 32.7 & 
scitail (tsv\_format)~\citep{scitail} & 56.3 & 48.6  \\
swedish\_reviews & 49.9 & 50.4 &
\makecell[l]{tamilmixsentiment \\ \citep{chakravarthi-etal-2020-corpus}} & 4.2 & 8.8 \\
told-br (binary)~\citep{DBLP:journals/corr/abs-2010-04543} & 53.7 & 46.2  &
\makecell[l]{tweet\_eval (emotion) \\ \citep{mohammad2018semeval}} & 39.4 & 59.3 \\

\bottomrule
\end{tabular}
\caption{Results of mgo-T5 and T5-3B on multi-lingual and multi-domain datasets.}
\label{tab:multilingual}
\end{table*}

\paragraph{Comparisons with T5 Variants.} Our key question  is whether the proposed method improves the zero-shot performance across various tasks.
 Table~\ref{tab:main} shows the results of the T5 variants and our method on public datasets. First, the top part of Table 3 shows the zero-shot results of the T5-base and T5-3B models. By comparing go-T5-3B and T5-3B, go-T5-base, and T5-base, we can conclude that \name is an effective method for improving zero-shot abilities. Even a small version, go-T5-small, can outperform T5-3B by a large margin.  Since go-T5 variants are fine-tuned on the specific self-supervised data, we also implement mgo-T5, a multi-task version of Go-tuning. It uses a single model for inference. We surprisingly observe that self-supervised data also can take advantage of task similarity. A small version, mgo-T5-base, achieves better results than go-T5-base. It demonstrates that multi-task pre-training can further improve the model's performance.

 \paragraph{Generalization to Downstream Tasks. } From Table~\ref{tab:main},  we also observe that self-supervised fine-tuning prefers ``natural'' tasks and shows large improvement variants across different tasks. As noted by~\citet{DBLP:journals/corr/abs-2005-14165, DBLP:conf/iclr/WeiBZGYLDDL22}, some tasks can be easily written into a natural language format, like sentiment classification and question answering, while some tasks not. For example, natural language inference is unlikely to appear in an unsupervised training set. For those ``natural'' tasks that are more likely to appear in raw texts, the proposed approach is a strong method that improves zero-shot performance by a large margin, such as from 64.1 to 75.4 in SST2, from 46.7 to 61.8 in BOOLQ, from 45.2 to 54.5 in PAWS.  For those ``unnatural'' tasks, the improvements of \name are somewhat marginal compared to the performance of ``natural'' tasks. On the other hand, these complicated tasks can be better handled by large language models, such as GPT-3. We hope that our work can encourage more research into how zero-shot capabilities emerge,  how LLMs learn such complicated abilities, and how smaller language models can learn complicated abilities from efficient self-supervised learning.

\paragraph{Comparisons with LLMs}  We also compare the proposed approach to the largest language models, including GPT-3 (175B)~\citep{DBLP:journals/corr/abs-2005-14165}, LaMDA-PT (137B)~\citep{DBLP:journals/corr/abs-2201-08239}, OPT (175B)~\citep{DBLP:journals/corr/abs-2205-01068}, and GLaM (64B)~\citep{DBLP:conf/icml/DuHDTLXKZYFZFBZ22}. These models are variants of GPT architectures. Following~\citet{DBLP:conf/iclr/WeiBZGYLDDL22}, we take the results of GPT-3 (175B), LaMDA-PT (137B), and GLaM (64B) from their respective papers. For OPT (175B), we re-implement OPT (175B) and use the same template in Table~\ref{tab:task} to get the results. Note that only OPT has a strictly fair comparison with the proposed method.  Table~\ref{tab:com_with_LLM} demonstrates that the proposed mgo-T5 model (250M), trained with a small number of self-supervised data, is on par with OPT models on the average performance. Furthermore, it outperforms GPT-3 in 3 out of 4 datasets (RTE, SST2, CB, and BOOLQ). However, it is important to note that despite with lower average score, large language models show strong performance on a subset of datasets, like SST2, BOOLQ. On these tasks, it is challenging to make smaller language models on par with LLMs only based on a small number of self-supervised data. In future work, we will focus more on these tasks to explore whether smaller language models can also learn zero-shot abilities as LLMs.

\paragraph{Ablation Studies} In this work, our objective is to explore how self-supervised fine-tuning help zero-shot abilities of language models. One key part is geometric-guided learning. We adopt the idea of kernel density estimation by considering the label distance to encourage the learning of a unified task, rather than learning multiple related tasks. We examine how performance is affected by removing geometric-guided learning. Table~\ref{tab:main} shows the results. The large improvement over T5-base-SDA reinforces our intuition that training on more data with the same direction matters.

\paragraph{Results on More Multilingual Tasks} In addition to the widely used GLUE and Super-GLUE benchmark, we also collect more datasets from huggingface\footnote{We will release all our used datasets, associated with templates and seed labels.}, covering multilingual datasets and multi-domain datasets.  We write a single prompt for each task. Table~\ref{tab:multilingual} shows the evaluation results. Although mgo-T5 is only fine-tuned on English-centric data, these results demonstrate that mgo-T5 can generalize to cross-lingual and cross-domain tasks. 

\section{Discussion}

Our paper aims to explore a fundamental question: can smaller language models have strong zero-shot abilities on unseen tasks? We operationalize this question by proposing a self-supervised fine-tuning method, Go-tuning, to fix the gap between a generalized landscape and a specific landscape. The proposed approach indeed greatly improves performance and even surpasses OPT-175B on the average performance. It means that self-supervised fine-tuning is indeed a key to improving smaller language models. Interestingly, the proposed approach, fine-tuned on English-centric data, can even be generalized to multi-lingual and multi-domain tasks. Both results show that self-supervised fine-tuning matters and ablation studies also show the effectiveness of the proposed geometric-guided learning.

However, we also notice that there are still some limitations. We find that large language models can have extremely high performance on a subset of datasets (generally complicated tasks), and such performance is hard to achieve by just adding a small number of self-supervised data. Future work on understanding the behavior of large models could be used to fix the gap between smaller language models and large language models.


\section{Conclusion}
This paper explores whether smaller language models can achieve strong zero-shot abilities as large language models do without any external supervision data. To achieve this goal, we proposed a simple but effective method for improving smaller language models. According to the proposed approach,  we build a small model mgo-T5 It achieves large performance improvements than the original T5 models and even reaches OPT-175B average performance over 9 tasks. Despite being promising, we also observe that there still remains a generalization gap between smaller language models and large language models on complicated tasks. 

\include{datasets_sts}

\bibliography{anthology,custom}
\bibliographystyle{acl_natbib}


\end{document}